\documentclass[11pt,a4paper]{article}

\usepackage[a4paper,top=0.8in,bottom=1.0in,left=0.8in,right=0.8in]{geometry}
\usepackage[utf8]{inputenc} 
\usepackage[T1]{fontenc}    
\usepackage[hidelinks]{hyperref}
\usepackage{url}            
\usepackage{booktabs}       
\usepackage{nicefrac}       
\usepackage{microtype}      
\usepackage{xcolor}         
\usepackage{graphicx}
\usepackage{subcaption}
\usepackage{cite}
\usepackage{multirow}
\usepackage{amsmath,amssymb,amsfonts}
\usepackage{bm}
\usepackage{wrapfig}
\usepackage{enumitem}
\usepackage{algorithmic}
\usepackage{algorithm}

\newcommand{\ie}{\textit{i.e.,}~}

\newcommand{\eg}{\textit{e.g.,}~}

\title{Knowledge Graph Fusion via Bidirectional Information Aggregation}

\author{
Songlin Zhai$^{1}$, Guilin Qi$^{1}$, Yue Wang$^{2}$, Yuan Meng$^{1}$ \\
{\small $^{1}$ Southeast University, Nanjing, China} \\
{\small $^{2}$ Stanford University, California, United States} \\
{\small \tt \{songlin\_zhai, gqi\}@seu.edu.cn, wyue0125@alumni.stanford.edu, yuan\_meng@seu.edu.cn}
}

\date{} 

\begin{document}

\maketitle

\begin{abstract}
  Knowledge graphs (KGs) play a critical role in enhancing large language models (LLMs) by introducing structured and grounded knowledge into the learning process. 
  However, most existing KG-enhanced approaches rely on parameter-intensive fine-tuning, which risks catastrophic forgetting and degrades the pretrained model's generalization.
  Moreover, they exhibit limited adaptability to real-time knowledge updates due to their static integration frameworks. 
  To address these issues, we introduce the first test-time KG-augmented framework for LLMs, built around a dedicated knowledge graph-guided attention (KGA) module that enables dynamic knowledge fusion without any parameter updates. 
  The proposed KGA module augments the standard self-attention mechanism with two synergistic pathways: outward and inward aggregation. 
  Specifically, the outward pathway dynamically integrates external knowledge into input representations via input-driven KG fusion. 
  This inward aggregation complements the outward pathway by refining input representations through KG-guided filtering, suppressing task-irrelevant signals and amplifying knowledge-relevant patterns. 
  Importantly, while the outward pathway handles knowledge fusion, the inward path selects the most relevant triples and feeds them back into the fusion process, forming a closed-loop enhancement mechanism. 
  By synergistically combining these two pathways, KGA supports real-time knowledge fusion exclusively at test-time, without any parameter modification. 
  Extensive experiments on five benchmarks verify the comparable knowledge fusion performance of KGA.
\end{abstract}

\section{Introduction}
The modern web has evolved into a vast information ecosystem, where the ability to access accurate, timely, and structured knowledge is paramount. 
As the cornerstone of the semantic web, large-scale Knowledge Graphs (KGs) like DBpedia \cite{mendes-etal-2012-dbpedia} and Wikidata \cite{10.1145/2629489} provide machine-readable facts that power a new generation of intelligent web applications, from semantic search engines \cite{10.1145/3511808.3557116,10.1145/3038912.3052558} to sophisticated recommendation systems \cite{10.1145/3696410.3714808,10.1145/3394486.3403143}. 
Concurrently, Large Language Models (LLMs) have emerged as the dominant paradigm for natural language interaction on the web. 
The synergy between these two technologies is therefore a critical frontier since knowledge graphs can provide structured symbolic knowledge to complement the implicit parametric knowledge encoded in large language models, giving rise to the emerging field of KG-enhanced LLMs \cite{saxena-etal-2020-improving,DBLP:journals/corr/abs-2107-02137,10.1145/3289600.3290956,DBLP:conf/iclr/JiangZ0W23,luo-etal-2024-chatkbqa}. 

\begin{figure}[htbp]
    \centering
    \includegraphics[page=1,width=0.6\linewidth]{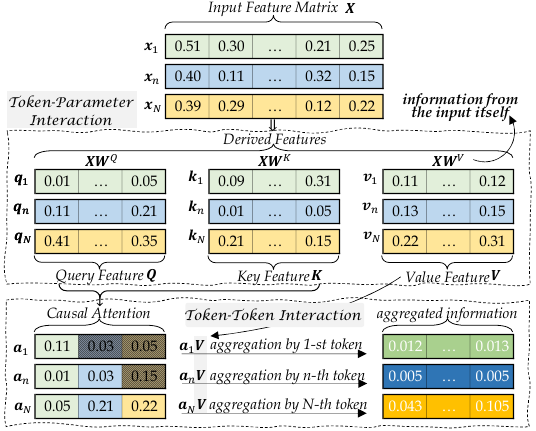}
    \caption{Motivation of \textit{why do we choose to rewire the ``self''-attention}, only expand part of the \textit{token-token} interaction.}
    \label{fig:attention-details}
\end{figure}

While existing methods demonstrate promise in grounding LLMs with structured facts, most of them predominantly rely on parameter-invasive fine-tuning strategies \cite{saxena-etal-2020-improving,DBLP:journals/corr/abs-2107-02137,10.1145/3289600.3290956,DBLP:conf/iclr/JiangZ0W23,kim-etal-2023-kg}. 
This fundamentally contradicts the preservation principle of LLMs \cite{wang2024wise}: parameter modification inevitably introduces catastrophic forgetting and degrades acquired capabilities. 
Furthermore, their static integration frameworks struggle to adapt to the real-time updates inherent in web-scale KGs. 
Alternative paradigms like Retrieval-Augmented Generation (RAG) \cite{huang-etal-2024-less,DBLP:journals/corr/abs-2405-12035} sidestep parameter updates but introduce new bottlenecks in retrieval reliability and latency \cite{DBLP:conf/nips/LewisPPPKGKLYR020}. 
%
Even recent advancements in long-context LLMs enable full-context knowledge injection \cite{li-etal-2024-retrieval,DBLP:journals/corr/abs-2409-01666,geminiteam2024geminifamilyhighlycapable,wang2025kblam}, yet they incur prohibitive computational overhead and suffer from an ``attention sink'' phenomenon, where information in the middle of long sequences is often ignored. 
These limitations motivate our central research question: 
\textit{\textbf{can we develop a parameter-preserving knowledge fusion method that achieves both efficient and real-time KG integration while maintaining an LLM's general capabilities?}} 

%
Drawing inspiration from \textit{Test-Time Adaptation} for LLMs \cite{zeng2025revisitingtesttimescalingo1like,muennighoff2025s1simpletesttimescaling,jurayj2025finalanswertesttimescaling} and the dual-pathway nature of human attention \cite{medicalattention2002}, this paper introduces the knowledge graph-guided attention (\textbf{KGA}), a novel inference-time framework that dynamically integrates external knowledge through non-invasively rewiring of the LLM's native attention mechanism. 
This design stems from a fundamental rethinking of the attention module in Transformer, \ie the self-attention mechanism inherently supports information aggregation through \textit{token-to-token} interactions \cite{DBLP:conf/iclr/CordonnierLJ20,DBLP:conf/acl/VoitaTMST19}. 
%
%
We thus reformulate the monolithic ``self''-aggregation of self-attention into a bidirectional \textit{``knowledge''-``text''} synergistic aggregation of KGA, 
achieved by introducing two complementary pathways that mimic human cognition: 
(1) \textbf{a fusion pathway}, which simulates a \textit{bottom-up} \textit{stimulus-driven} process in human brain. It dynamically integrates external knowledge into the input representation via the rewired attention. 
\textit{Just as humans, when confronted with a question, our brain uses its semantic cues to trigger the integration of relevant information from external sources} \cite{medicalattention2002}. 
(2) \textbf{a guidance pathway}, which simulates a \textit{top-down} goal-directed verification. It assesses the contextual importance of each candidate knowledge triple. 
\textit{Similarly, once a tentative answer emerges, humans instinctively revisit the original question to seek evidence that supports or refutes it} \cite{medicalattention2002}.

Specifically, the \textit{bottom-up} pathway performs the initial input-driven fusion of all candidate KG triples based on the \textit{query-key} interactions within the attention. 
Concurrently, the \textit{top-down} pathway calculates a relevance score for each triple by using it to query the input text, effectively measuring how much evidence the input provides for that piece of knowledge. 
These scores are then normalized into a set of adaptive weights. 
Crucially, these two pathways work in synergy. The adaptive weights from the \textit{top-down} guidance pathway are used to modulate the aggregation process in the \textit{bottom-up} fusion pathway. 
This allows KGA to perform an adaptive weighted fusion of external knowledge, amplifying the signals from the most relevant triples while suppressing noise from irrelevant ones. 
This entire mechanism reuses the LLM's native attention weights since LLM itself has already demonstrated its ability to understand knowledge \cite{wang2025kblam,DBLP:journals/kbs/DaiHWSJQ25}. 
Our design preserves architectural integrity and treating KGs as a plug-and-play component for real-time updates without any parameter modification. 

\noindent Extensive experiments on four benchmarks demonstrate the effectiveness and efficiency of our proposed framework. 
To summarize, the contributions of this paper could be listed as follows: 
%
\begin{itemize}[leftmargin=1.5em]
    \item \textbf{Cognitively Inspired Attention Rewiring}: A novel method that elegantly rewires self-attention to mimic the dual \textit{bottom-up} and \textit{top-down} pathways of human cognition, enabling a more lightweight knowledge integration for LLMs. 
    \item \textbf{Adaptive Knowledge Fusion}: A parameter-free, inference-time mechanism that supports adaptive fusion of external knowledge, guided by a dynamic, context-aware relevance assessment. 
    \item \textbf{Comprehensive Validation}: Extensive experiments across four datasets on diverse tasks, demonstrating KGA's superior performance and efficiency for real-time knowledge fusion. 
\end{itemize}



\section{Related Work}

\subsection{Knowledge Enhancement for LLMs}
 
Large language models exhibit impressive generative and reasoning capabilities, but still struggle with factual consistency and structured reasoning. 
To address these limitations, researchers augment LLMs with structured knowledge graphs, either through training-time integration or inference-time augmentation. 
\textbf{Training-time enhancement} injects KGs during the pretraining or fine-tuning phase. 
For example, K-BERT \cite{DBLP:conf/aaai/LiuZ0WJD020} augments the input text with entities from a knowledge graph during pretraining, while ERNIE \cite{DBLP:journals/corr/abs-2107-02137} and CoLAKE \cite{DBLP:conf/coling/SunSQGHHZ20} employ unified pretraining objectives to integrates KGs, enhancing LLMs' ability to leverage structured knowledge. 
Additionally, Kformer \cite{DBLP:conf/nlpcc/YaoHDWCZ22} injects knowledge into the feedforward layer, providing a simple and effective alternative to attention-based integration. 
KnowLA \cite{luo-etal-2024-knowla} has begun to leverage these adapter-based PEFT strategies for knowledge injection. 
These methods have shown success in improving the model's understanding of world knowledge, but are inherently static, relying on fixed training data. 
%
\textbf{Inference-time methods} augment LLMs with external knowledge during decoding, making them more flexible. 
%
For example, joint reasoning over KG subgraphs \cite{DBLP:conf/naacl/YasunagaRBLL21} integrates external knowledge through subgraph retrieval, which enables more structured reasoning by querying relevant parts of the knowledge graph during inference. 
Multi-stage retrieval and inference pipelines \cite{DBLP:conf/emnlp/KimKJC23} combine external knowledge retrieval with multi-step reasoning, improving the accuracy of model outputs by progressively refining the information used at each step. 
Non-parametric memory fusion for language generation \cite{DBLP:conf/nips/LewisPPPKGKLYR020} uses external knowledge stored in a memory bank, dynamically fusing it with LLMs during the generation process, without requiring additional fine-tuning. 
Parallel context windows \cite{DBLP:conf/acl/RatnerLBRMAKSLS23} enable LLMs to process multiple context windows simultaneously, improving modularity in inference. 
KELP \cite{DBLP:conf/acl/LiuWZDL24}, also explore latent semantic matching to enhance the model's ability to select the most relevant paths in the knowledge graph. 
Recent exemplars include G-Retriever \cite{he2024gretriever}, which introduces a RAG framework for general textual graphs, and RAGRAPH \cite{jiang2024ragraph} further refine RAG frameworks for graph-based data. Concurrently, frameworks facilitating iterative reasoning enable LLMs to perform multi-step exploration over the KG, such as Think-on-Graph \cite{DBLP:conf/iclr/SunXTW0GNSG24} and iQUEST \cite{wang-yu-2025-iquest}, leading to more robust predictions.

\subsection{Differences between KGA and GAT}
It is important note that our proposed KGA is conceptually and functionally distinct from \textit{graph attention networks} (GATs) \cite{velikovi2018graph}, despite the similarity in nomenclature. 
GATs are a class of neural networks designed to learn representations on graph structures by introducing a trainable attention mechanism that assigns weights of node's neighbors. 
In contrast, KGA is a parameter-free mechanism for fusing external knowledge into a large language model via its native attention architecture. 
It operates on textualized knowledge triples, not graph topology, and re-purposes existing self-attention module to achieve adaptive fusion of external facts. 

\subsection{Challenges \& Positioning This Work}
Despite significant advancements, most existing methods (whether training-time or inference-time) rely on parameter-intensive training or complex inference pipelines to improve knowledge retrieval. 
They typically struggle with fine-grained control over the knowledge fusion process and are not ideally suited for applications requiring low latency and high adaptability. 
\textit{Our work offers a distinct perspective within the landscape of KG-enhanced LLMs}. Inspired the cognitive dual-pathway architecture, KGA achieves non-invasive, inference-time integration by merely rewiring the internal self-attention module. 
This design positions KGA as a lightweight, efficient, and highly adaptable solution for integrating real-time, dynamic web knowledge into LLMs, while ensuring the soundness and reliability of the fusion process. 

\section{Knowledge Graph-Guided Attention}

\subsection{Notations and Task Definition}
%

Let $\mathcal{M}$ denote a pre-trained large language model that receives a textual prompt $X$ as input. 
The input $X$ is tokenized into a sequence of tokens $X=\{x_1,..., x_n,..., x_N\}$ where $N$ is the sequence length. 
At each transformer layer $l\in[1,L]$, the hidden representation of token $x_n$ is denoted as the bold item $\bm{x}_n^{(l)}$, with $\bm{X}^{(l)} = [\bm{x}_1^{(l)};...;\bm{x}_n^{(l)}...;\bm{x}_N^{(l)}]$ representing the full input feature matrix at layer $l$. 
For notational simplicity, the layer superscript is omitted ($\bm{x}_n, \bm{X}$) when discussing operations within a single layer. 

External knowledge $\mathcal{G}$ is provided as a set of triples; we use $Z \in \mathcal{G}$ to denote a specific triple in the form of (\textit{subject, relation, object}), \eg (\textit{WWW\_2026, held\_in, Dubai}). 
Our goal is to develop an integration mechanism that could dynamically incorporate these triples into the forward computation of $\mathcal{M}$ to enhance its ability to generate the desired output $Y^*$ in response to the input $X$. 

\begin{figure}[tb]
    \centering
    \includegraphics[page=1,width=0.5\linewidth]{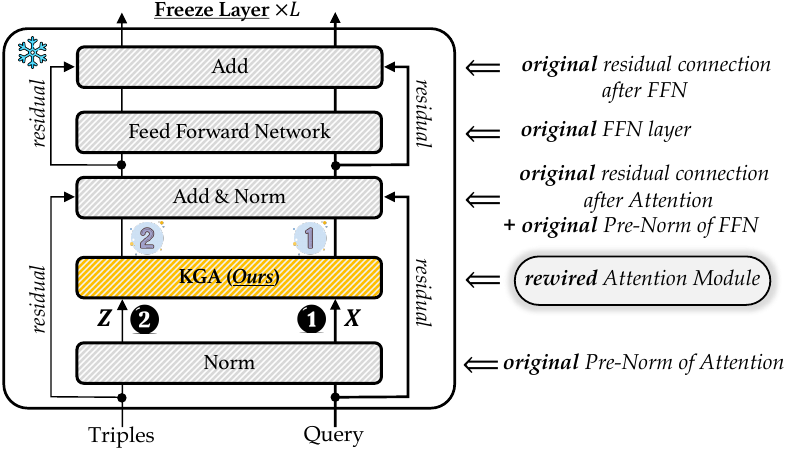}
    \caption{Hidden Transformer layer equipped with KGA.}
    \label{fig:architecture}
\end{figure}

\subsection{Cognitive Inspiration \& Overall Framework}
As illustrated in Figure~\ref{fig:attention-details}, we exploit the core functionality of the Transformer's self-attention. Since this module inherently supports flexible \textit{token–token} interactions, simply rewiring the attention is sufficient to integrate external knowledge rather than training other new components. 
This leads to the central design challenge: \textit{how can this rewiring be accomplished in a principled and effective manner?}
To address this, we draw inspiration from cognitive neuroscience \cite{medicalattention2002}, which suggests that humans employ two synergistic attention mechanisms when solving problems requiring background knowledge. 
The first is a \textit{bottom-up} information integration process, driven by the salience of external stimuli. 
For instance, when faced with a question, the semantic cues within the problem itself trigger our brain to search for relevant information from external sources. 
This stimulus-driven process treats the question as the stimulus signal. 
\textbf{In KGA, the input query serves as this stimulus signal.} 
The second is a \textit{top-down} verification process, guided by internal goals or expectations. For example, when having a tentative answer in mind, we automatically revisit the original question to find evidence that supports or refutes it. 
This is a goal-directed process, with the verification of the answer acting as the goal. 
\textbf{In KGA, the goal corresponds to quantifying the importance of each triple.} 
Our framework is explicitly designed to mimic this powerful dual-process system. 
Figure~\ref{fig:architecture} shows the overall architecture of each transformer layer equipped with KGA, where the only difference is the substitution of the vanilla self-attention with KGA. 

\begin{figure*}[t]
    \centering
    \includegraphics[page=1,width=0.99\textwidth]{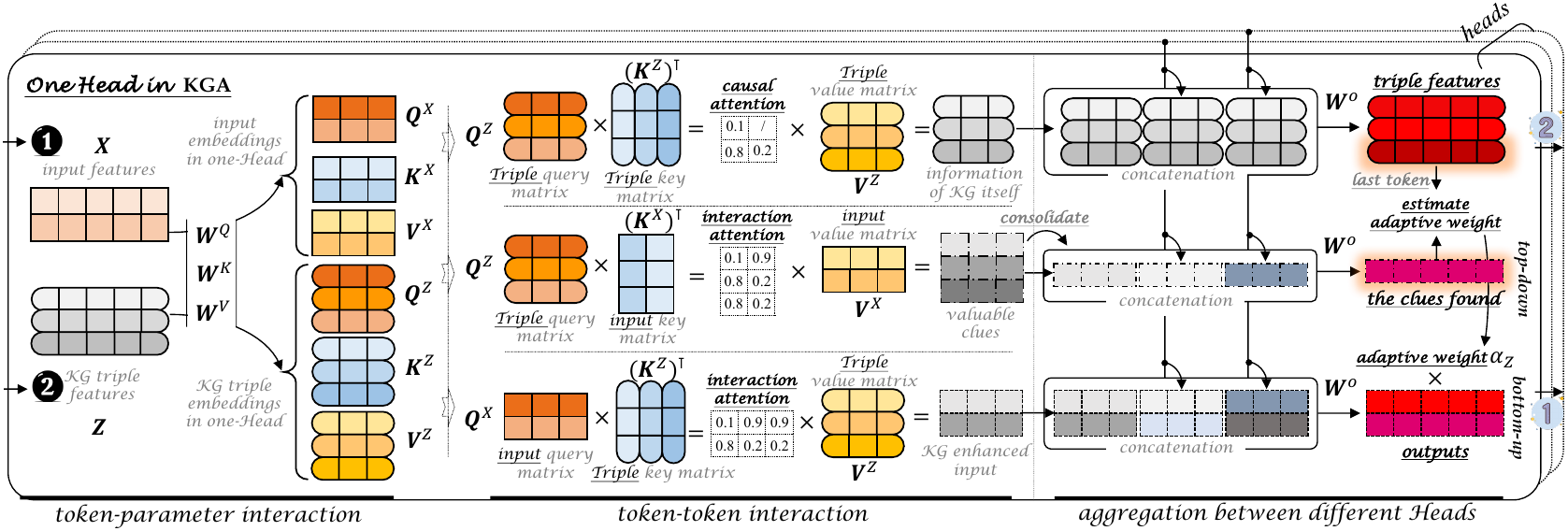}
    \caption{Framework of the proposed KGA, where the \textit{bottom-up} and \textit{top-down} pathways are introduced in it.}
    \label{fig:kga}
\end{figure*}

Figure~\ref{fig:kga} illustrates the details of the proposed KGA. 
Specifically, these two pathways operate based on the information flows between the input $X$ and external knowledge $Z\in\mathcal{G}$.
For the \textit{bottom-up} pathway, the \textit{query} matrix of input $X$ is used to probe the \textit{key} matrix of knowledge triples $Z$, aggregating the corresponding information from the \textit{value} matrix of $Z$, thereby achieving knowledge fusion. 
The \textit{top-down} guidance pathway operates analogously but in reverse. It employs the \textit{query} matrix of the triple $Z$ to probe the \textit{key} matrix of the input $X$, aggregating valuable clues from the input \textit{value} matrix. 
These clues are then used to estimate the importance of each knowledge triple, which drives the adaptive weighting process in the integration step. 
The details of these pathways will be elaborated in the following sections.

\subsection{\textit{Bottom-Up}: Knowledge Fusion Pathway}\label{sec:flow1}
The primary function of this pathway is to dynamically integrate external information through input-driven knowledge aggregation, emulating the \textit{bottom-up} stimulus-driven nature of human attention. 
For a given triple $Z$, it is integrated into the input features layer by layer. 
Formally, the triple hidden representations are first transformed into \textit{query}, \textit{key} and \textit{value} features: 
\begin{equation}
    \bm{Q}^Z = \bm{Z} \bm{W}^Q, \quad
    \bm{K}^Z = \bm{Z} \bm{W}^K, \quad
    \bm{V}^Z = \bm{Z} \bm{W}^V
\label{eq:kg-embed}
\end{equation}
where $\bm{Z}=[\bm{z}_1,\allowbreak ...,\allowbreak \bm{z}_m,\allowbreak ...,\allowbreak \bm{z}_M]$ is the feature matrix for a triple with $M$ tokens. 
The derived components $\bm{q}^Z_m \in \bm{Q}^Z$, $\bm{k}^Z_m \in \bm{K}^Z$ and $\bm{v}^Z_m \in \bm{V}^Z$ denote the \textit{query}, \textit{key} and \textit{value} features of the $m$-th triple token in $Z$, respectively. 
Crucially, by using the original $\bm{W}^Q$, $\bm{W}^K$ and $\bm{W}^V$ matrices in the self-attention module, we ensure that the knowledge triples are projected into the same feature space as the input, enabling meaningful semantic interaction without introducing new parameters. 
%
The updated representation $\hat{\bm{x}}_n$ is then formulated as:
%
%
\begin{equation}
\begin{aligned}
    %
    %
    \hat{\bm{x}}_n = 
    \underbrace{
    \sum_{i=1}^N \frac{\mathrm{Exp}(w^X_{n,i}) \cdot \bm{v}_i^X}{\sum_{j=1}^N \mathrm{Exp}(w^X_{n,j})}
    }_{\text{\textit{aggregation: input itself}}} + 
    \underbrace{
    \sum_{Z\in\mathcal{G}} \alpha_Z \sum_{m=1}^M \frac{\mathrm{Exp}(r^Z_{n,m}) \cdot \bm{v}^Z_m}{\sum_{j=1}^M \mathrm{Exp}(r^Z_{n,j}) / \sqrt{D}}
    }_{\text{\textit{aggregation: external knowledge}}}
\end{aligned}
\label{eq:stream2}
\end{equation}
where $w^X_{n,i} = \{\bm{q}^X_n (\bm{k}^X_i)^{\top}\}/\sqrt{D}$ is the original attention score of input to itself. 
$r^Z_{n,m}=\{\bm{q}^X_n (\bm{k}^Z_m)^{\top}\}/\sqrt{D}$  is the attention score between input token $x_n$ and triple token $z_m$. 
$\alpha_Z$ is the adaptive importance of the entire triple $Z$, which is computed by the \textit{top-down} guidance pathway (detailed in Section~\ref{sec:flow2}).

%
%
This design enables input representations to actively ``pull'' information from external knowledge though the simple non-invasive token-token interactions. 
It offers three key advantages: contextual adaptability (knowledge influence dynamically adjusts to input semantics), architectural purity (no interference with existing model architecture), and computational efficiency (linear scaling with triple number). 
This fusion operates systematically across all transformer layers, allowing for hierarchical specialization, \ie lexical-level fusion occurring in lower layers, and semantic-level integration taking in higher layers. 
%


\subsection{\textit{Top-Down}: Attention Guidance Pathway}\label{sec:flow2}
Complementing the \textit{bottom-up} fusion, this pathway emulates a \textit{top-down}, goal-directed cognitive process. 
Its primary function is to quantify the contextual importance of each knowledge triple $Z$, thereby computing the adaptive weight $\alpha_Z$, used in Eq.~\ref{eq:stream2}. 
This is achieved through a three-stage process: \textit{clue extraction}, \textit{consolidation}, and \textit{relevance scoring}. 

\subsubsection{Clue Extraction} 
First, this pathway performs a KG-guided refinement of the input representation to extract valuable ``clues''. This process uses each knowledge triple as a lens to focus on relevant parts of the input, strategically suppressing noise while amplifying knowledge-critical patterns. This helps address potential attention dispersion issues seen in standard self-attention \cite{ahmed2025mateiclmitigatingattentiondispersion,ye2025differential}. 
Formally, for each token $z_m$ in a triple $Z$, we compute a clue vector $\bm{c}_m$ by using its query representation $\bm{q}^Z_m$ to attend to the entire input $\bm{X}$: 
\begin{equation}
    \bm{c}_m = \underbrace{
    \sum^{N}_{i=1}\bigg\{
    \text{Exp}(\frac{\bm{q}^Z_m (\bm{k}^X_i)^{\top}}{\sqrt{D}}) \bm{v}^X_i
    \bigg\}/\bigg\{
    \sum^{N}_{j=1} \text{Exp}(\frac{\bm{q}^Z_m (\bm{k}^X_j)^{\top}}{\sqrt{D}})
    \bigg\}
    }_{\text{\textit{extracting a clue from $X$ guided by triple token $z_m$}}}
\label{eq:stream3}
\end{equation}
where $\bm{c}_m$ is the resulting clue vector. This non-causal attention mechanism allows for full visibility of the input context, ensuring a comprehensive recalibration. The resulting matrix $\bm{C} = \{ \bm{c}_1,\allowbreak ...,\allowbreak \bm{c}_m,\allowbreak ...,\allowbreak \bm{c}_M\}$ thus encodes a set of clue vectors from input, each refined and viewed from the perspective of a different token in the triple $Z$. 

\subsubsection{Clue Consolidation}
Next, we consolidate the \textit{token-level} clue vectors \{$\bm{c}_1$...$\bm{c}_m$\} into a single, \textit{triple-level} representation. A simple summation would be suboptimal, as not all tokens in a triple contribute equally to its meaning, potentially diluting critical information. We therefore propose an attention-aware merging strategy that leverages the triple's own internal semantics: 
\begin{equation}
    \hat{\bm{c}} = \bm{w}^{Z}_{M} \bm{C} = \sum\nolimits_{m=1}^M w^Z_{M,m} \bm{c}_m
\label{eq:summarization}
\end{equation}
Here, $w^Z_{M,m}$ is the causal self-attention between the $M$-th triple token (the last one) and the $m$-th token in $Z$. This design cleverly re-purposes the LLM's inherent understanding of token importance within the triple to guide the consolidation process. 

\subsubsection{Relevance Scoring}
Finally, the consolidated clue representation $\hat{\bm{c}}$ is used to quantifies the triple's contextual relevance to the input $X$. This is formulated by a dot product between $\hat{\bm{c}}$ and the hidden representation of the last triple token $\bm{z}_M$: 
\begin{equation}
    s(Z, X) = \hat{\bm{c}} \ (\bm{z}_M)^{\top}
\label{eq:importance}
\end{equation}
This scalar $s(Z, X)$ provides a dynamic, context-aware measure of the triple's relevance. 
After computing the raw relevance scores for each candidate triple in the set $\mathcal{G}$, we normalize them into a probability distribution using a temperature-scaled softmax. 
The temperature ($\tau$) is a hyperparameter that controls the sharpness of the final weight distribution: 
\begin{equation}
    \alpha_Z = \frac{\mathrm{Exp}(s(Z, X) / \tau)}{\sum_{Z' \in \mathcal{G}} \mathrm{Exp}(s(Z', X) / \tau)}
\label{eq:adaptive-weighting}
\end{equation}
Here, $\alpha_Z$ represents the final, normalized importance weight for the triple $Z$. 
These weights are then used directly in the \textit{bottom-up} fusion pathway (as seen in Eq.~\ref{eq:stream2}) to modulate the contribution of each knowledge triple. 
This temperature scaling provides a mechanism to control the model's behavior: a higher $\tau$ creates a softer distribution, encouraging exploration of more triples, while a lower $\tau$ creates a sharper distribution, forcing the model to focus confidently on only the most relevant facts. This temperature scaling provides a flexible mechanism to balance selectivity (focusing on highly relevant knowledge) and diversity (exploring a broader set of potentially useful facts) during knowledge integration.

\section{Experiments}


%
%
\begin{table}[tb]
\tabcolsep=4.5pt
\centering
\caption{Comparison between KGA and previous methods on different tasks, where methods with $^{\dag}$ are the ones based on GPT-3.5 and the ones with $^{\star}$ are based on DeepSeek-V3.}
\begin{tabular}{|lc|lr|}
%
%
%
%
%
\multicolumn{2}{c}{\textbf{\textit{(KGQA)}}} &\multicolumn{2}{c}{\textbf{\textit{(KG Reasoning)}}} \\
\midrule[0.1pt]
\multicolumn{2}{|c|}{\textbf{{SimpleQuestions}}} &\multicolumn{2}{|c|}{\textbf{{PathQuestion}} (\textit{2-hop})} \\
\hline
\textbf{\textit{Methods}} &{\textit{Hit@1}} &\textbf{\textit{Methods}} &{\textit{Hit@1}} \\
\hline
MemNN \cite{DBLP:journals/corr/BordesUCW15} &63.9 &ISM \cite{DBLP:conf/icdm/LanW019} &99.1 \\
CFO \cite{DBLP:conf/acl/DaiLX16} &62.6 &QAGCN \cite{DBLP:conf/esws/WangRCB24} &98.5 \\
AMPCNN \cite{DBLP:conf/coling/YinYXZS16} &67.2 &Uhop-HR \cite{chen-etal-2019-uhop} &97.6 \\
Character \cite{DBLP:conf/emnlp/HeG16} &70.3 &KV-MemNN \cite{miller-etal-2016-key} &97.4 \\
IOPrompt \cite{DBLP:conf/nips/BrownMRSKDNSSAA20} &20.0 &SRN \cite{10.1145/3336191.3371812} &96.3 \\
SC \cite{DBLP:conf/iclr/0002WSLCNCZ23} &18.9 &IRN \cite{zhou-etal-2018-interpretable} &96.0 \\
RoG \cite{DBLP:conf/iclr/LuoLHP24} &73.3 &MINERVA \cite{DBLP:conf/iclr/DasDZVDKSM18} &75.9 \\
StructGPT$^{\dag}$ \cite{DBLP:conf/emnlp/JiangZDYZW23} &50.2 &TransferNet \cite{shi-etal-2021-transfernet} &93.2 \\
PoG$^{\star}$ \cite{DBLP:conf/nips/ChenTJSYX24} &63.9 &RL-MHR \cite{DBLP:conf/icml/DasGNTZHJM22} &94.1 \\
KnowPath$^{\star}$ \cite{DBLP:journals/corr/abs-2502-12029} &65.3 &AlAgha \cite{chen-etal-2019-uhop} &97.4 \\
ToG$^{\star}$ \cite{DBLP:conf/iclr/SunXTW0GNSG24} &59.7 &ARN$\_{\text{DistMult}}$ &84.3 \\
\hline
\textbf{ZSL}   (Qwen2.5 \textbf{0.5B}) &{19.5} &{ZSL} (Qwen2.5 \textbf{0.5B}) &{20.1} \\
\textbf{ICL}   (Qwen2.5 \textbf{0.5B}) &{67.3} &{ICL} (Qwen2.5 \textbf{0.5B}) &{73.7} \\
\textbf{KGA}   (Qwen2.5 \textbf{0.5B}) &\textbf{75.1} &\textbf{KGA} (Qwen2.5 \textbf{0.5B}) &{98.3} \\
\midrule[0.3pt]
\end{tabular}
\label{tab:kgqa-results}
\end{table}

\subsection{Experimental Setup}\label{sec:exp:setting}
\subsubsection{Experimental Settings}
To validate the effectiveness of our knowledge fusion framework, we conduct experiments across three distinct tasks: knowledge graph question answering (KGQA), knowledge graph reasoning, and knowledge-based model editing (KME). 
For KGQA and KG reasoning tasks, models are required to integrate knowledge triples from external KGs to answer the given questions. While naively feeding the entire KG to the model is computationally prohibitive, we also aim to avoid reliance on complex retrieval systems that could confound our results. Instead, to create a set of candidate triples, we first perform entity linking on the question mentions, then retrieve all 1-hop or 2-hop neighboring triples as candidates (\ie $\mathcal{G}$). 
While this method of selecting candidate triples may introduce noise triples, it creates a realistic testbed for validating the robustness of our fusion method against imperfectly retrieved knowledge, a scenario that closely mirrors real-world web applications. 
Complementarily, the KME task aims to rectify the outdated or error knowledge in the model by using external KGs. 
We focus on the challenging yet practical scenario of \textit{consecutive editing}, where the model must assimilate a sequence of new facts constantly. 
This setup directly simulates a real-world web environment where a knowledge-aware application must continuously integrate a stream of updates from an evolving KG to maintain its factuality. 
\textit{It serves as an ideal testbed for evaluating the model's adaptability to the dynamic nature of knowledge, a central theme of our work.} 
All experiments of all methods are conducted on an \textit{Linux NVIDIA A100 80GB} ($256$ GB RAM) machine.

\subsubsection{Baselines}
For comparative evaluation in KGQA and KG Reasoning, we compare against four categories of baselines: (1) \textbf{Supervised Fine-Tuned Methods (SFT)}: representing traditional supervised approaches with KG-augmented training, 
(2) \textbf{In-Context Learning (ICL)}: flattening all candidate KB triples into a natural language utterance and then concatenating it in front of the input question.  
(3) \textbf{Zero-Shot Learning (ZSL)}: directly processing input questions without KG integration to establish reference performance levels of merely using the LLM's internal knowledge for answering. 
(4) \textbf{other advanced inference-time} methods, including RoG, PoG and ToG. 
%
For knowledge-based model editing, we compare KGA against a strong suite of specialized baselines in this task, including finetuning (FT-C, LoRA), parameter-editing (\eg ROME, AlphaEdit), and memory-based methods (GRACE, WISE). 

\begin{table*}[tb]
\tabcolsep=10pt
\centering
\caption{Comparison between KGA and prevailing methods under $3K$ \textit{consecutive editing}. (a dynamic knowledge environment)}
\begin{tabular}{l|ccc|ccr}
\toprule[1.0pt]
\multicolumn{1}{l}{\multirow{2}{*}{\textbf{Methods}}} &\multicolumn{3}{c}{\textbf{\textsc{CounterFact}} \cite{DBLP:conf/nips/MengBAB22}} &\multicolumn{3}{c}{\textbf{\textsc{ZsRE}} \cite{DBLP:conf/conll/LevySCZ17}} \\
\cline{2-7}
\multicolumn{1}{c}{} &\small{\textit{Efficacy}} &\small{\textit{Generality}} &\multicolumn{1}{c}{\small{\textit{Locality}}} &\small{\textit{Efficacy}} &\small{\textit{Generality}} &\small{\textit{Locality}} \\
\midrule[0.5pt]
Llama3 (8B) (ZSL) &0.0087 &0.0075 &/  &0.2627 &0.2598 &/ \\
\midrule[0.5pt]
FT-C \cite{DBLP:conf/nips/MengBAB22} &0.0575 &0.0047 &0.0013   &0.0769 &0.0666 &0.0069  \\ 
LoRA \cite{DBLP:conf/iclr/XuXG0CZC0024} &0.0077 &0.0117 &0.0017  &0.1145 &0.1116 &0.0535     \\ 
ROME \cite{DBLP:conf/nips/MengBAB22} &0.2507 &0.1323 &0.0097   &0.0339 &0.0280 &0.0015        \\ 
R-ROME \cite{gupta-etal-2024-rebuilding} &0.4892 &0.3662 &0.0147  &0.0271 &0.0243 &0.0035     \\ 
MEMIT \cite{DBLP:conf/iclr/MengSABB23} &0.0000 &0.0000 &0.0722  &0.0000 &0.0000 &0.0396       \\ 
AlphaEdit \cite{DBLP:journals/corr/abs-2410-02355} &0.0033 &0.0017 &0.0007  &0.0001 &0.0000 &0.0003   \\ 
PMET \cite{DBLP:conf/aaai/Li0SYMY24} &0.0000 &0.0000 &0.0000    &0.0000 &0.0000 &0.0000  \\
EMMET \cite{DBLP:conf/emnlp/GuptaSA24} &0.0517 &0.0486 &0.0043  &0.5450 &0.3882 &0.0128  \\
GRACE \cite{DBLP:conf/nips/HartvigsenSPKG23} &0.0003 &0.0000 &0.9938   &0.0624 &0.0095 &1.0000   \\ 
WISE \cite{wang2024wise} &0.1473 &0.0763 &0.9907      &0.3348 &0.3283 &0.9997      \\ 
\midrule[0.5pt]
IKE (ICL) \cite{DBLP:conf/emnlp/ZhengLDFWXC23} &0.0055 &0.0043 &0.6509    &0.5233 &0.5231 &0.5289  \\ 
\textbf{KGA} &\textbf{0.6485} &\textbf{0.3963} &\textbf{1.0000}  &\textbf{0.5912} &\textbf{0.5629} &\textbf{1.0000}  \\ 
\midrule[0.5pt]
\midrule[0.5pt]
Qwen2.5 (7B) (ZSL)     &0.0078 &0.1340 &/   &0.2016 &0.1942 &/ \\
\midrule[0.5pt]
FT-C \cite{DBLP:conf/nips/MengBAB22}       &0.0407 &0.0150 &0.0003    &0.0239 &0.0199 &0.0029\\ 
LoRA \cite{DBLP:conf/iclr/XuXG0CZC0024}      &0.0200 &0.0200 &0.0020   &0.0527 &0.0516 &0.0072 \\ 
ROME \cite{DBLP:conf/nips/MengBAB22}       &0.0000 &0.0000 &0.0242    &0.2809 &0.2538 &0.0924 \\ 
R-ROME \cite{gupta-etal-2024-rebuilding}    &0.5778 &0.1742 &0.4585   &0.5910 &0.5016 &0.3871  \\ 
MEMIT \cite{DBLP:conf/iclr/MengSABB23}      &0.0000 &0.0000 &0.0000   &0.0000 &0.0000 &0.0000  \\ 
AlphaEdit \cite{DBLP:journals/corr/abs-2410-02355}   &0.0000 &0.0000 &0.0000       &0.0001 &0.0001 &0.0002  \\ 
PMET \cite{DBLP:conf/aaai/Li0SYMY24}   &0.0002 &0.0001 &0.0000    &0.0003 &0.0001 &0.0000  \\
EMMET \cite{DBLP:conf/emnlp/GuptaSA24} &0.0700 &0.0604 &0.0054  &0.5705 &0.4323 &0.0281  \\
GRACE \cite{DBLP:conf/nips/HartvigsenSPKG23}        &0.0012 &0.0000 &0.9939  &0.2905 &0.0095 &1.0000  \\ 
WISE \cite{wang2024wise}      &0.0352 &0.0283 &0.0645        &0.2747 &0.2690 &0.0985\\ 
\midrule[0.5pt]
IKE (ICL) \cite{DBLP:conf/emnlp/ZhengLDFWXC23}    &0.5001 &0.2272 &0.5686    &0.6974 &0.6889 &0.5029   \\ 
\textbf{KGA} &\textbf{0.6408} &\textbf{0.4772} &\textbf{1.0000} &\textbf{0.8374} &\textbf{0.8000} &\textbf{1.0000}  \\ 
\bottomrule[1.0pt]
\hline
\end{tabular}
\label{tab:editing-results}
\end{table*}

\begin{figure*}[tb]
     \centering
     %
     \includegraphics[page=1,width=0.24\textwidth]{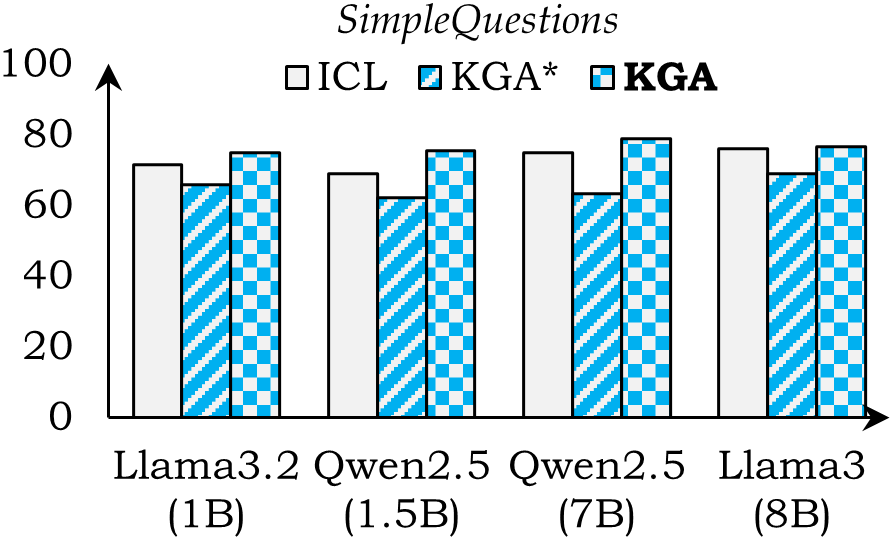}
     \includegraphics[page=1,width=0.24\textwidth]{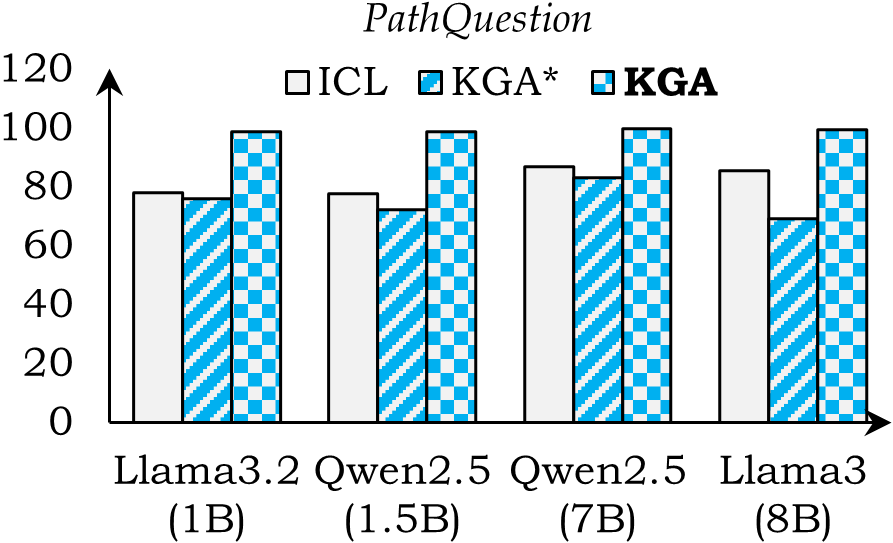}
     \includegraphics[page=1,width=0.24\textwidth]{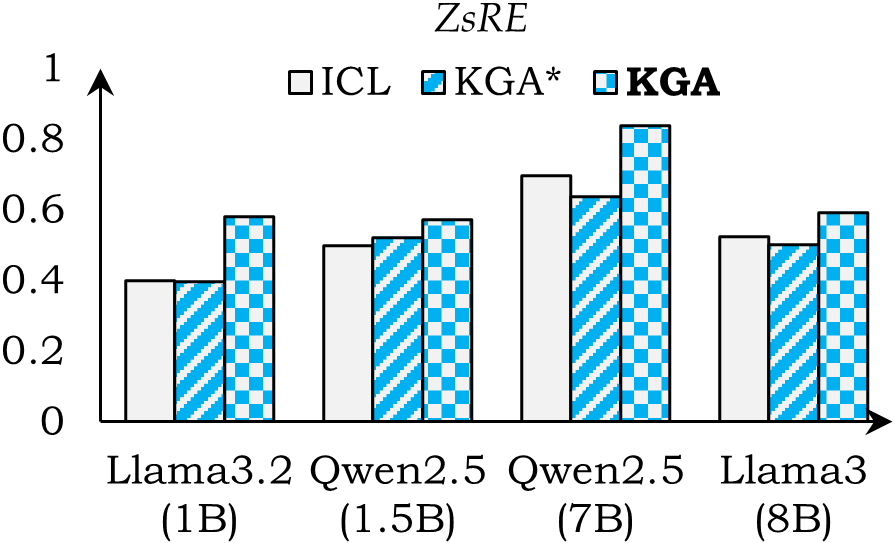}
     \includegraphics[page=1,width=0.24\textwidth]{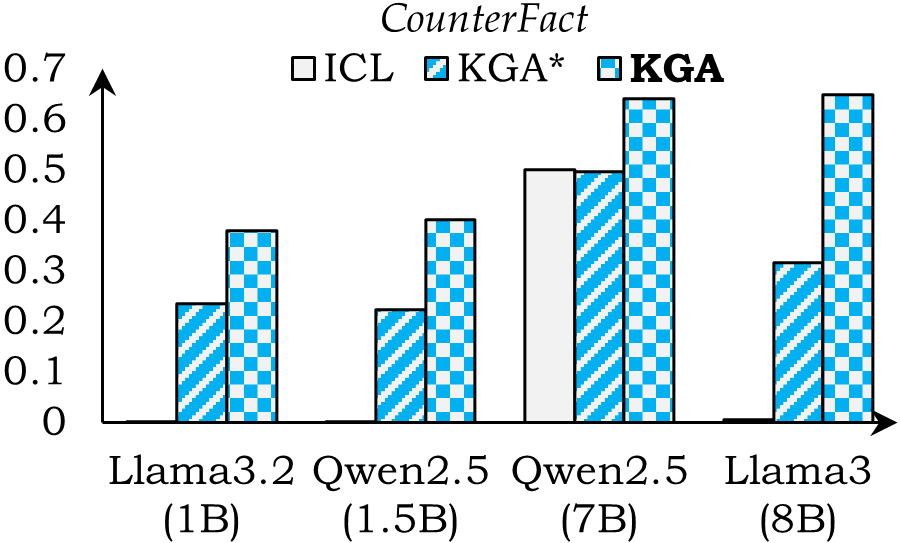}
    \caption{Effects of adaptive weighting on four datasets, where KGA* denotes the version of KGA without \textit{top-down} pathway.}
    \label{fig:adaptive-weighting}
\end{figure*}

\subsection{Overall Comparison}
\subsubsection{Performance on KGQA and KG Reasoning}
Table~\ref{tab:kgqa-results} provides a comprehensive performance comparison on KGQA and KG reasoning benchmarks, revealing three critical insights. 
First, \textbf{SFT}-based models achieve impressive, even near-perfect in-domain results (\eg ISM attains $>99\%$ accuracy on PathQuestion \textit{2-Hop}). 
While this type of models fit the current data well, extensive studies reveal that such models' practical applicability is limited by severe well-documented drawbacks: \textbf{poor cross-domain transferability} \cite{luo-etal-2024-chatkbqa} (\eg $40^+\%$ accuracy drop in Wikidata$\to$DBpedia transfers), \textbf{weak adaptability to KG dynamics} \cite{trung-etal-2024-reft}, and \textbf{inability to support incremental learning}. 
Second, ICL significantly improves accuracy over the zero-shot baseline (\eg boosting PathQuestion \textit{2-Hop} from $25.1\%$ to $73.7\%$) but introduces prohibitive computational costs and memory overhead (detailed in Sections~\ref{sec:exp:efficiency} and~\ref{sec:exp:memory}). 
This inefficiency stems from its indiscriminate aggregation of all candidate triples. 
By feeding the entire triple set into the self-attention mechanism without explicit weighting mechanisms, ICL inevitably suffers from redundant computations over irrelevant triples (\textbf{suboptimal efficiency}) and \textbf{degraded interpretability} due to uncontrolled cross-triple token interactions (see Appendix~\ref{appendix:sec:dis}) that obscure knowledge utilization patterns \cite{wang2025kblam}. 
These drawbacks underscore the need for a more targeted fusion method. Third, our KGA demonstrates a compelling balance of performance and practicality. It consistently outperforms the ICL baseline through its adaptive, fine-grained knowledge fusion (\eg $20^+\%$ performance improvement on PathQuestion \textit{2-Hop}). Remarkably, this high performance is achieved even when KGA is applied to a very small model (Qwen2.5 0.5B), yielding results that are competitive with, and in some cases not inferior to, heavily fine-tuned SFT methods. 
This highlights the efficiency of our attention-rewiring approach. Furthermore, KGA is highly adaptable, enables real-time KG updates, and preserves the model's general capabilities (evidenced by $100\%$ \textit{Locality} in KME). The adaptive mechanism also provides quantifiable interpretability by tracing layer-wise attention (Section~\ref{sec:exp:case-study}). These results position KGA as a powerful and practical solution for deployments requiring both knowledge awareness and computational efficiency. 
Crucially, our framework remains decoupled from retrieval filtering mechanisms, enabling seamless integration with advanced retrieval techniques like dense semantic search (\eg DPR \cite{karpukhin-etal-2020-dense}) or cross-encoder reranking to further boost model performance. 


\begin{figure*}[tb]
     \centering
     %
     \includegraphics[page=1,width=0.24\textwidth]{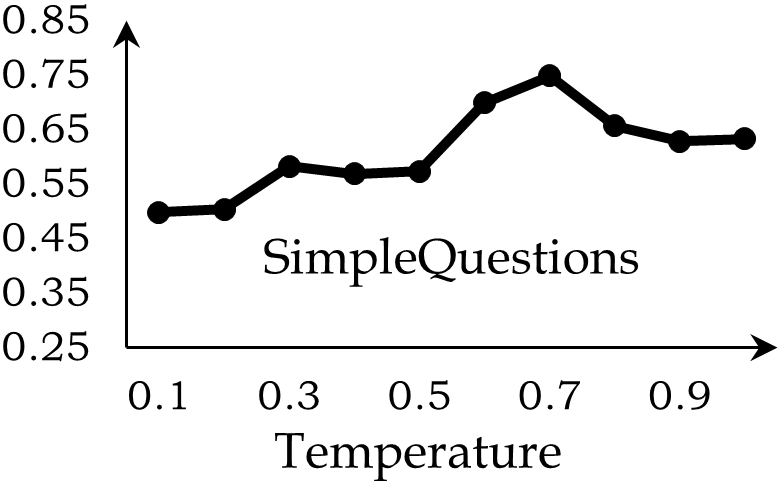}
     \includegraphics[page=1,width=0.24\textwidth]{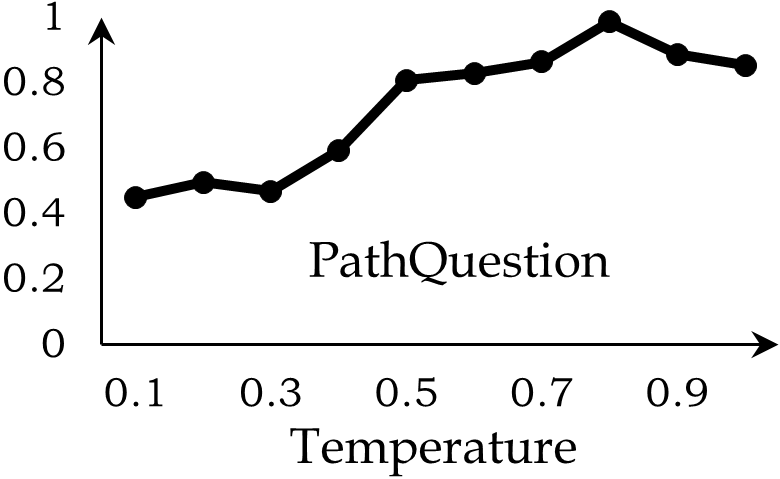}
     \includegraphics[page=1,width=0.24\textwidth]{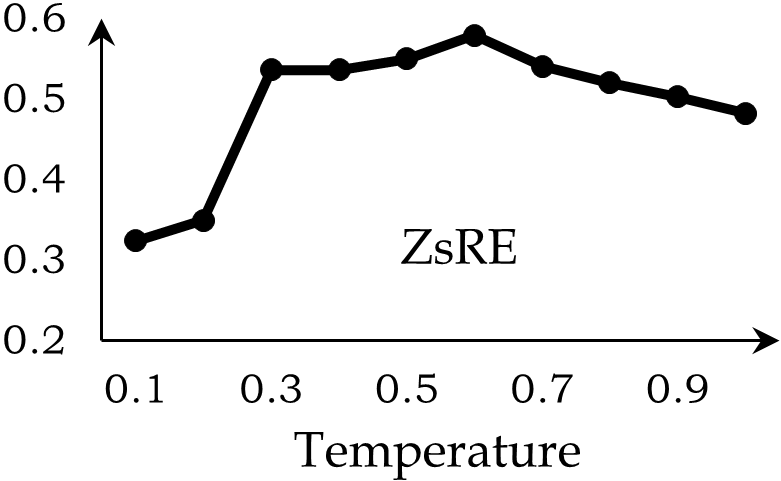}
     \includegraphics[page=1,width=0.24\textwidth]{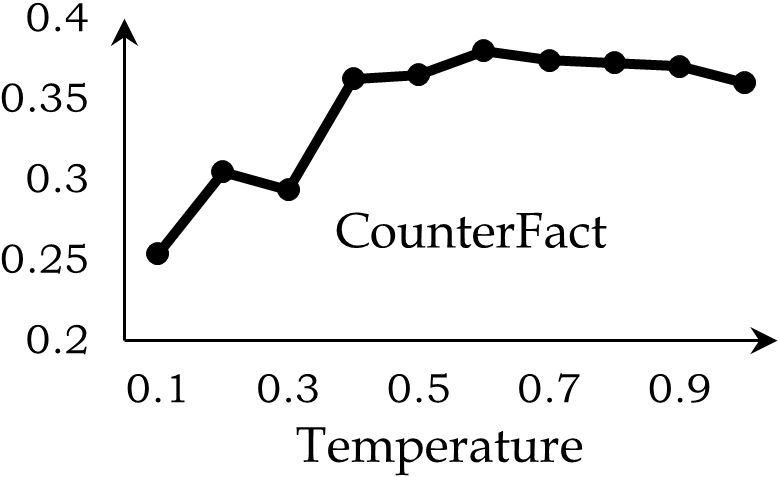}
     \caption{Analysis of \textit{Top-Down} Pathway: effects of temperature on four datasets for Llama3.2 (1B).}
    \label{fig:temperature}
\end{figure*}

\subsubsection{Performance on Knowledge Editing}
Table~\ref{tab:editing-results} presents the primary results for KME on \textsc{ZsRE} and \textsc{CounterFact} using three standard metrics in this task: 
\textit{Efficacy} (adherence to the external knowledge), \textit{Generality} (generalization to paraphrased queries), and \textit{Locality} (retention of unrelated parametric knowledge). 
The candidate triples are retrieved from constantly accumulated knowledge using BM25. 
Generally, the results reveal a significant challenge for existing methods in adapting to evolving knowledge of real-world applications. 
The ``\textit{Original}'' rows quantify the base models' pre-existing knowledge, indicating the lack of new knowledge (\ie analogous to the \textbf{ZSL} baseline). 
While standard In-Context Knowledge Editing (IKE) attempts to provide the external knowledge as context information, it struggles with fidelity, with Llama3 (8B) achieving $0.55\%$ \textit{Efficacy} on \textsc{CounterFact}. 
This failure highlights the inadequacy of simple context-stuffing and underscores the need for a more robust integration mechanism. 
In stark contrast, our KGA demonstrates remarkable effectiveness. It boosts the \textit{Efficacy} of Llama3 (8B) on \textsc{CounterFact} from $0.55\%$ to $65.45\%$, a more than $100$-fold improvement. This substantial gain, achieved without any parameter updates, showcases KGA's ability to reliably integrate new facts, making it a highly effective solution for keeping LLMs aligned with real-time knowledge.

\begin{figure*}[tb]
     \centering
     %
     \includegraphics[page=1,width=0.24\textwidth]{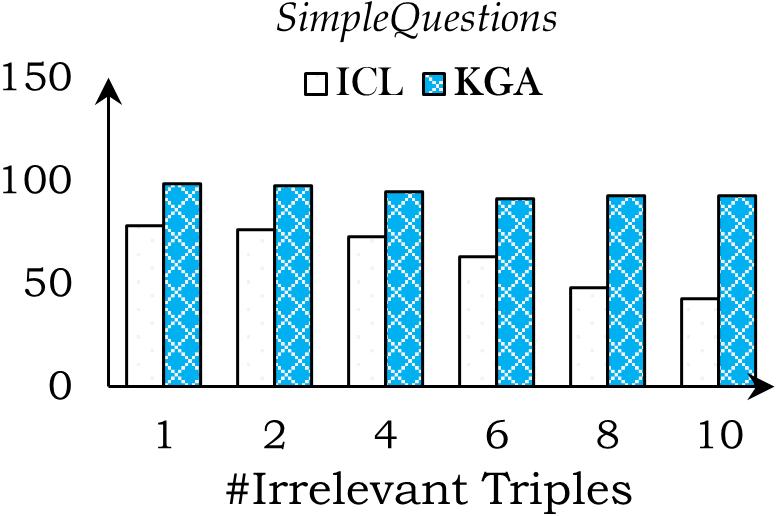}
     \includegraphics[page=1,width=0.24\textwidth]{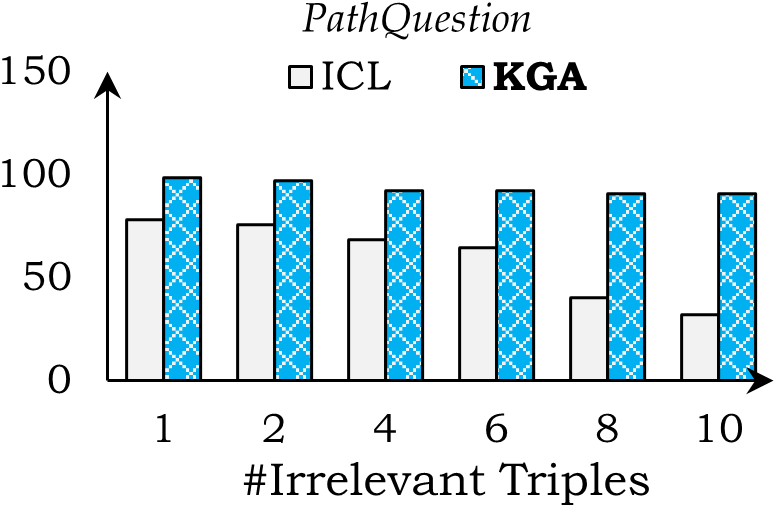}
     \includegraphics[page=1,width=0.24\textwidth]{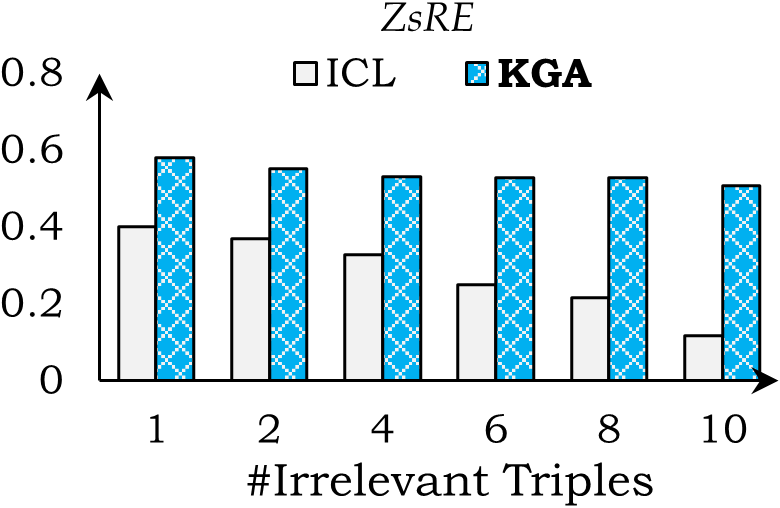}
     \includegraphics[page=1,width=0.24\textwidth]{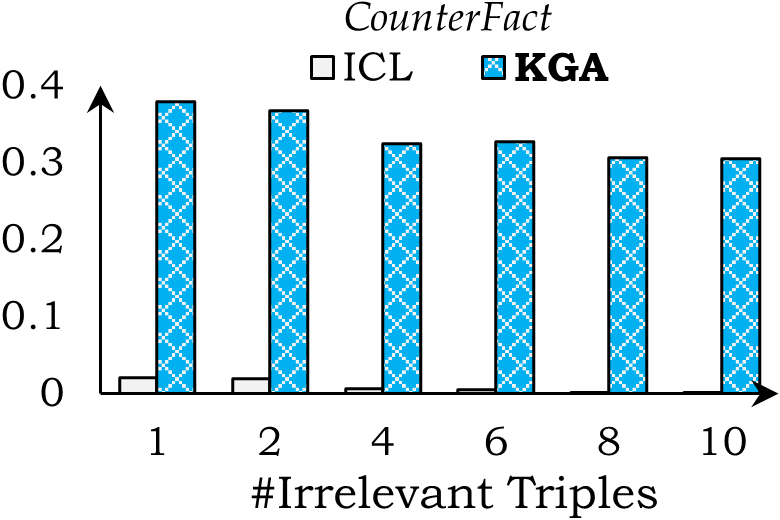}
    \caption{Analysis of \textit{Top-Down} Pathway: robustness to noise on four datasets for Llama3.2 (1B).}
    \label{fig:noise}
\end{figure*}

\subsection{Analysis of \textit{Top-Down} Pathway}
Inspired by the \textit{top-down} nature of human attention, our framework quantifies the contextual relevance of each knowledge triple. This adaptive weighting mechanism not only enhances KGA's effectiveness but also improves its tolerance to noisy data. In this section, we analyze this component in detail. 
\subsubsection{Effects of Adaptive Weighting}
To evaluate the effectiveness of this pathway, we perform an ablation by replacing the adaptive weighted fusion in Eq.~\ref{eq:stream2} with a naive summation. 
In this variant, denoted as KGA*, all candidate triples are assigned identical importance (\ie $1$). 
Figure~\ref{fig:adaptive-weighting} compares the performance of KGA* against the full KGA framework and the ICL baseline across various datasets and models. 
The results show a substantial performance degradation when adaptive weighting is removed. On several datasets, the performance of KGA* even drops below that of ICL. 
This stark decline underscores the critical role of adaptive weighting: simply fusing knowledge is insufficient and the ability to dynamically assess and prioritize knowledge is essential. 

\subsubsection{Effects of Temperature in Adaptive Weighting}
When computing adaptive weights $\alpha_Z$, we introduce a temperature $\tau$ to control the sharpness of the triple relevance distribution. 
A lower temperature makes the distribution ``sharper'', causing the model to concentrate on only the highest-scoring triples. Conversely, a higher temperature ``softens'' the distribution, encouraging the model to consider a broader set of triples. 
To assess the impact of this hyperparameter, we conducted experiments on four datasets. Figure~\ref{fig:temperature} illustrates the performance of Llama3.2 as a function of $\tau$. 
Results reveal a consistent trend: performance tends to increase as the temperature rises from a low value, peaking at a moderate level before gradually declining as the temperature becomes too high, which can overly flatten the distribution and introduce noise. \textit{Although temperature has an influence, the performance variation is not drastic, indicating that KGA is not overly sensitive to it}. \textbf{It is also worth noting that the temperature is the sole hyperparameter in KGA, highlighting its simplicity and ease of practical use.} 

\subsubsection{Effects of Noise Suppression}
By assessing triple relevance through a \textit{top-down} verification process, KGA can effectively mitigate the negative impact of irrelevant triples on the fusion process, an advantage that ICL lacks. To stress-test this robustness, we evaluate performance under increasing proportions of noisy (irrelevant) triples. 
Figure~\ref{fig:noise} compares KGA and ICL as the noise ratio grows. 
While ICL's performance degrades sharply, KGA remains remarkably stable. 
This resilience to noise is crucial in real-world web applications, where retrieved knowledge is inevitably imperfect. 
KGA's ability to handle noise demonstrates its practical viability.

\subsection{Efficiency Comparison}\label{sec:exp:efficiency}
\begin{figure}[tb]
    \centering
    \includegraphics[page=1,width=0.5\linewidth]{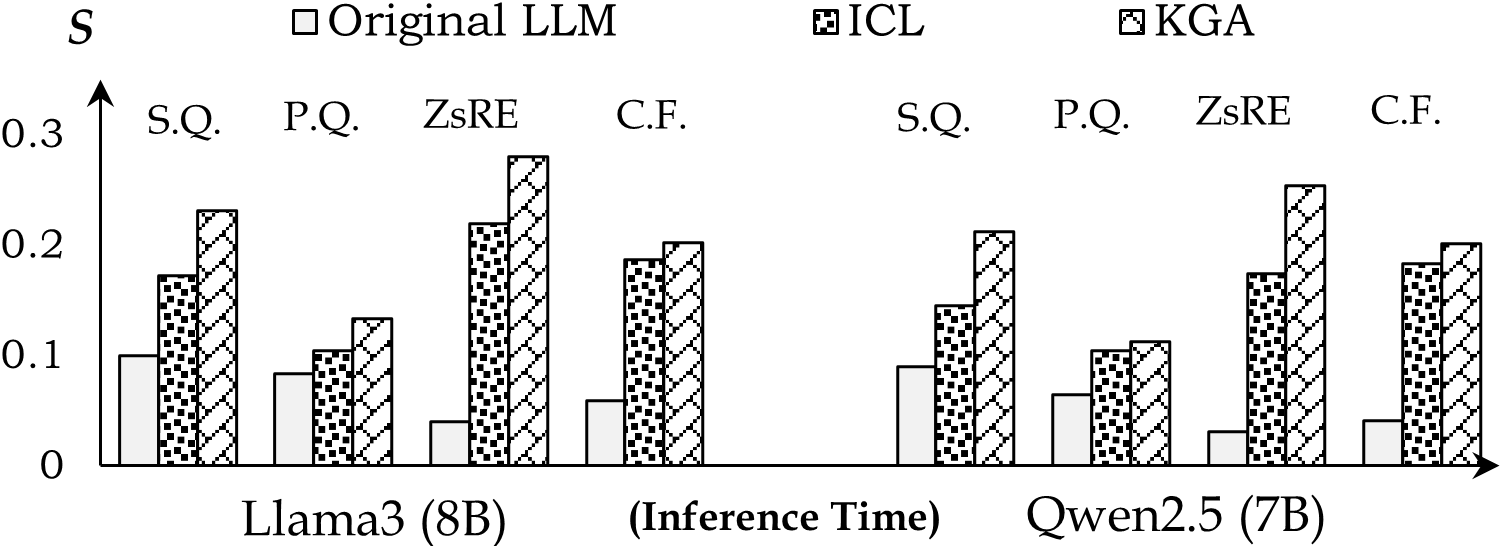}
    \caption{Efficiency comparison of three different methods.}
    \label{fig:efficiency}
\end{figure}

%
Our framework leverages external knowledge to recalibrate input representations for valuable clue extraction, thereby quantifying the contextual relevance of each triple for knowledge integration.  
This estimation process brings about additional computation. 
In practical applications, inference efficiency is a critically important indicator. 
Hence, we measure latency across three methods under four datasets, shown in Figure~\ref{fig:efficiency}. 
The ZSL baseline maintains consistently low latency ($0.1$ \textit{s/instance} on PathQuestion) as it processes no triples. 
In contrast, ICL exhibits high inference time, \eg processing $100$ triples incurs about $20\times$ higher latency than ZSL due to indiscriminate attention over all candidate triples. 
Moreover, ICL exhibits quadratic growth in inference time with respect to the number of triples. 
A large portion of this computation is wasted on irrelevant triples interaction in self-attention that are not useful for the current query (see Appendix~\ref{appendix:sec:dis}). 
By contrast, KGA introduces a marginal computational overhead for relevance estimation while achieving substantial performance gains. 
\textit{Importantly, these measurements represent the full cost of our method, including the parallel pathway of adaptive weight estimation, without any optimizations such as caching or pre-computation}. 
These results demonstrate that KGA is an inherently lightweight mechanism that can be integrated into existing inference pipelines with negligible impact, confirming its practicality for real-world web application.

\begin{figure}[tb]
    \centering
    \includegraphics[width=0.5\linewidth]{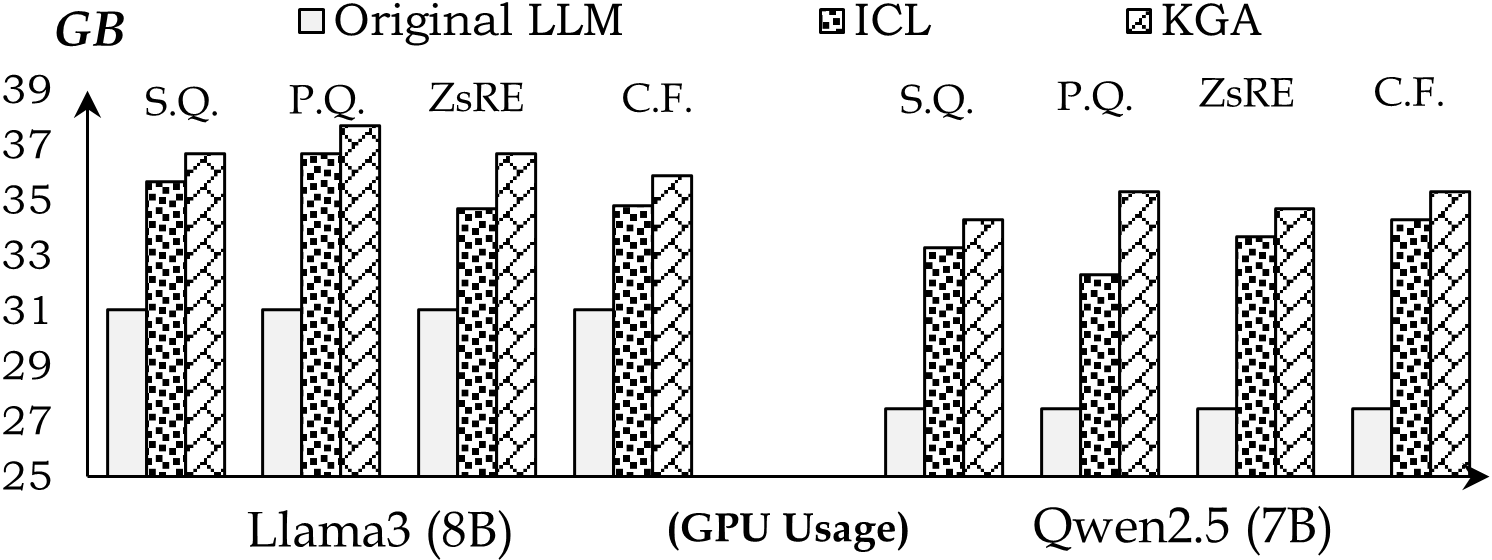}
    \caption{GPU memory comparison of different methods.}
    \label{fig:memory-usage}
\end{figure}


\subsection{Memory Usage}\label{sec:exp:memory}
%

Memory footprint constitutes another critical deployment consideration for knowledge fusion systems. 
We evaluate GPU memory consumption across three methods under four datasets, benchmarking against the base model's requirement. 
The ZSL baseline maintains stable memory usage ($\sim 33$ GB for Llama3 (8B)) as it processes no triples, while ICL exhibits dramatic memory growth, requiring $78$ GB when feeding $100$ triples. 
The memory requirement of ICL will further escalate for token-dense triples (\eg $112^+$GB for $100$ verbose triples). 
This overhead stems largely from caching non-critical triple features and maintaining dense token-level interactions across all triples (as analyzed in Appendix~\ref{appendix:sec:dis}). 
However, our method addresses this by cutting off the unnecessary interactions between different triples, thereby dramatically reducing the memory requirement during attention computation. 
At the same time, this pruning also helps mitigate the negative influence of overly long triple sequences on model performance. 
This stability enables deployment on resource-constrained devices, positioning KGA as a practical solution for memory-sensitive knowledge applications. 

\subsection{Case Study}\label{sec:exp:case-study}
\begin{figure}[tb]
    \centering
    \includegraphics[width=0.6\linewidth]{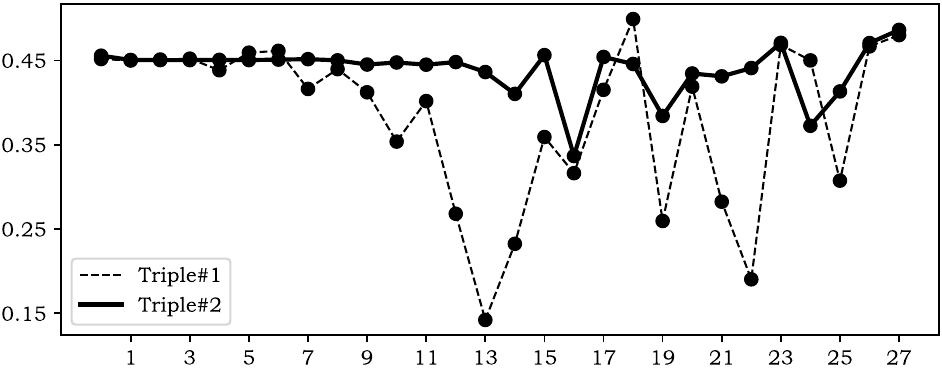}
    \caption{Triple relevance across different hidden layers of Qwen2.5 (7B), where Triple\#1 refers to \textit{(WWW\_2026, held\_in, Dubai)}, and Triple\#2 represents \textit{(Dubai, located\_in, UAE)}, with the query being \textit{In which country is WWW\_2026 held?}.}
    \label{fig:case-study}
\end{figure}

To better illustrate how KGA utilizes input triples and also clarify its interpretability, we analyze a concrete example: answering the question ``\textit{In which country is WWW\_2026 held?}'' with two input triples: (\textit{WWW\_2025, held\_in, Dubai}) and (\textit{Dubai, located\_in, UAE}). 
This is a two-hop reasoning case requiring synthesis of both triples. 
Figure~\ref{fig:case-study} shows the layer-wise triple relevance allocated to each triple (computed using Eq.~\ref{eq:importance}). 
We observe the following progression: 
\textbf{In shallow layers (1–4)}: both triples receive comparable attention as the model focuses on low-level linguistic processing (\eg entity recognition and syntactic parsing). 
\textbf{in intermediate layers (4–7)}: slightly higher attention is assigned to the first triple due to its direct mention of the query entity \textit{WWW\_2026}. 
\textbf{In deeper layers (8–17)}: attention shifts strongly to the second triple, reflecting its critical role in completing the geographic reasoning chain (\textit{located\_in}$\to$\textit{country}).
\textbf{For final layers (18–28)}: oscillating attention patterns emerge, indicating iterative refinement and cross-triple evidence integration during answer generation. 
This \textit{hierarchical attention progression} aligns with the characteristic processing of LLMs, \ie transitioning from shallow feature grounding $\to$ intermediate semantic association $\to$ deep relational reasoning $\to$ final prediction synthesis \cite{DBLP:journals/corr/abs-2003-08271,DBLP:conf/coling/DingCD0022}. 
Such interpretability highlights KGA's ability to selectively emphasize the most relevant triples at different abstraction levels, further supporting its explainability.

\section{Conclusion}\label{sec:conclusion}

This paper introduced knowledge graph-guided attention (KGA), a novel inference-time knowledge fusion framework that dynamically integrates external knowledge into large language models. 
By rewiring the native self-attention mechanism, KGA avoided any parameter modification while preserving the model's pre-trained capabilities. 
Inspired by the dual-pathway nature of human cognition, our method established a synergistic interaction between a \textit{bottom-up} fusion pathway and a \textit{top-down} guidance pathway. 
This design enables principled, adaptively weighted knowledge fusion, setting KGA apart from prior methods that rely on simple context augmentation or fine-tuning. 
Extensive experiments demonstrated KGA's performance, efficiency and robustness to noise.  Its parameter-free and adaptive fusion design makes KGA a practical solution for deploying reliable, knowledge-aware LLMs that can seamlessly adapt to the dynamic, real-time nature of web-scale knowledge graphs.

\bibliography{main}


\section{Discussion of Difference between KGA and Other Methods}\label{appendix:sec:dis}
\begin{figure*}[tb]
     \centering
     \begin{subfigure}[b]{0.325\textwidth}
         \centering
         \includegraphics[page=1,width=\textwidth]{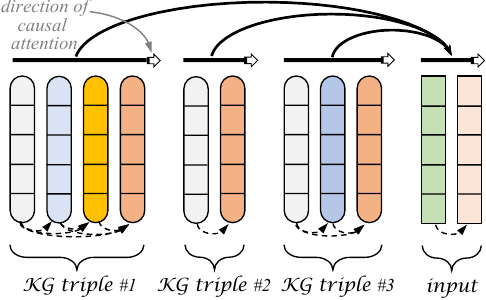}
         \caption{The proposed KGA.}
         \label{fig:kga-demo}
     \end{subfigure}
     \hfill
     \begin{subfigure}[b]{0.325\textwidth}
         \centering
         \includegraphics[page=1,width=\textwidth]{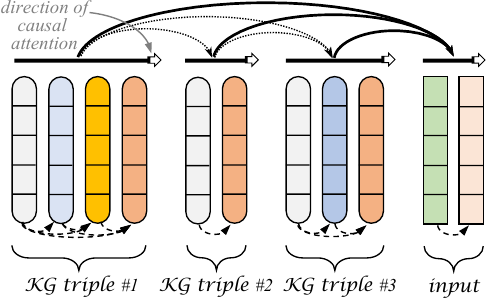}
         \caption{In-Context Learning.}
         \label{fig:icl-demo}
     \end{subfigure}
     \begin{subfigure}[b]{0.325\textwidth}
         \centering
         \includegraphics[page=1,width=\textwidth]{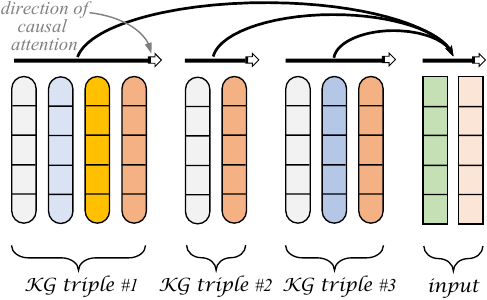}
         \caption{Traditional Cross-Attention.}
         \label{fig:cross-demo}
     \end{subfigure}
    \caption{Illustrative depiction of information aggregation under different methods.}
    \label{fig:demo}
\end{figure*}

Section~\ref{sec:flow1} delineates the external knowledge enhancement mechanism through input-driven triple integration (\ie \textit{bottom-up} aggregation). 
While this pathway shares superficial similarities with several conventional approaches, our framework exhibits fundamental architectural innovations. 
As shown in Figure~\ref{fig:kga-demo}, the proposed pathway systematically integrates three essential fusion dimensions: (1) Triple-to-Input: Knowledge infusion from KG to input sequence, (2) Triple-Itself: Intra-token self-attention for semantic consolidation in a given triple (3) Input-Itself: Native self-attention for contextual coherence maintenance. 
This tripartite architecture enables simultaneous refinement of both input representations and knowledge embeddings through their intrinsic patterns. 

However, \textbf{In-Context Learning} in Figure~\ref{fig:icl-demo} introduces uncontrolled inter-triple interference due to concatenating all triples as extended input. 
This forced co-processing of potentially unrelated triples (a) induces interference of different triples, (b) introduces sensitivity to triple ordering, and (c) unnecessary computation of inter-triple interaction and decreases the model efficiency. 
\textbf{Cross-Attention} in Figure~\ref{fig:cross-demo} restricts to unidirectional knowledge fusion (Triple$\to$Input) while failing to preserve essential attention flows of the input itself (\ie the first term in Eq.~\ref{eq:stream2}). 
This architectural deficiency prevents effective knowledge refinement (no Triple-Itself) and degrades input contextualization (without Input-Itself). 

\newcommand{\algcomment}[1]{{\small\textsf{\color{gray}#1}}}
\begin{algorithm}[htbp]
\caption{Process of language models equipped with KGA}
\label{appendix:alg:kga}
\begin{algorithmic}[1]
\REQUIRE Candidate triples $\mathcal{G}$, Query $X$, temperature $\tau$.
\ENSURE Model predictions $\mathcal{M}_{\bm{\Theta}}(X)$ based on our KGA.
\STATE \algcomment{// Initialize representations of $Z \in \mathcal{G}$, $X$ from token embeddings}
\STATE $\bm{Z}^{(0)}, \bm{X}^{(0)} \gets \texttt{TokenizeAndEmbed}(Z, X)$
\STATE \algcomment{// Model Forward Pass}
\FOR{\textit{layer\_idx} in \texttt{enumerate}(\textit{hidden\_layers})}
\STATE \algcomment{// 1. Estimate triple relevance}
\STATE \algcomment{// 1.1. Extract valuable clues}
\STATE Estimate $\bm{C}$ via Eq.~\ref{eq:stream3} based on $\bm{X}$ and $\bm{Z}$
\STATE \algcomment{// 1.2 Perform forward pass of triples}
\STATE $\bm{Z}^{(l+1)}\gets$ \texttt{KGA\_layer}($\bm{Z}^{(l)}$)
\STATE \algcomment{// 1.3 Consolidate clues and compute relevance}
\STATE Consolidate and perform $\alpha_Z$ estimation via Eq.~\ref{eq:summarization} - Eq.~\ref{eq:adaptive-weighting}
\STATE \algcomment{// 2 Adaptive integration}
\STATE Perform knowledge fusion via Eq.\ref{eq:stream2}
\STATE $\bm{X}^{(l+1)}\gets$ \texttt{KGA\_layer}($\bm{X}^{(l)}$)
\ENDFOR
\STATE $\text{Logits} \gets \texttt{LMHead}(\bm{X}^{(L)})$
\RETURN $\texttt{Decode}(\text{Logits})$
\end{algorithmic}
\end{algorithm}

\end{document}